\documentclass[letterpaper, 10pt, journal]{IEEEtran}
\IEEEoverridecommandlockouts
\usepackage{amssymb, amsmath}
\usepackage{xcolor, xcolor-solarized}
\usepackage{todonotes}
\usepackage[T1]{fontenc}
\usepackage[switch]{lineno}
\usepackage{booktabs}
\usepackage{multirow}
\usepackage{tikz}
\usepackage{siunitx}
\usepackage{makecell}
\usepackage{url}
\usepackage{ifthen}
\usepackage[htt]{hyphenat}
\usepackage{colortbl}
\usepackage{pifont}

\usetikzlibrary{arrows.meta, positioning, calc, fit, backgrounds, decorations.pathreplacing, matrix}

\definecolor{paperblue}{HTML}{2196F3}
\input{keyidea}
\def\outlinemode{\clean}

\newcommand{\cmark}{\textcolor{green!60!black}{\ding{51}}}
\newcommand{\xmark}{\textcolor{red!80!black}{\ding{55}}}

\newcommand{\R}{\mathbb{R}}
\newcommand{\M}{\mathcal{M}}
\newcommand{\Prior}{\mathcal{P}}
\newcommand{\I}{\mathcal{I}}
\newcommand{\name}{LOCI}

\title{Leveraging Semantic Maps for City-Scale Cross-View Localization}

\author{
  Ethan Fahnestock$^{*1}$ \qquad Erick Fuentes$^{*1}$ \qquad Philip R Osteen$^2$ \qquad Nicholas Roy$^1$\\
  $^1$ CSAIL, MIT $^2$ DEVCOM Army Research Laboratory (ARL) 
\thanks{$^*$ indicates equal contribution. Distribution Statement A: Approved for public release: distribution unlimited.}
\thanks{This work has been submitted to the IEEE for possible publication. Copyright may be transferred without notice, after which this version may no longer be accessible.}
}

\begin{document}
\maketitle
\begin{abstract}
\begin{keyidea}[minlevel=annotated]{X}
We want robots to localize in previously untraversed environments against commonly available prior data.
Rich semantic data available from OpenStreetMap can be useful in this task.
However, existing methods either ignore this semantic information, directly matching panoramas and overhead imagery, or dramatically compress the semantic information, working with a small set of fixed classes.
\end{keyidea}
\begin{keyidea}[minlevel=annotated]{Y}
 \,To leverage this rich semantic information, two challenges need to be overcome.
 First, useful semantic information needs to be extracted from the robot's egocentric observations.
 Second, the observed information must be quickly associated with the large prior semantic map (e.g., up to 628\,km$^\mathbf{2}$).
\end{keyidea}
\begin{keyidea}[minlevel=annotated]{Z}
\,We show that VLMs are effective at both extracting relevant landmarks from panoramas, and identifying feasible correspondences between these landmarks and prior overhead landmarks.
However, using VLMs to propose all correspondences scales poorly as the number of mapped landmarks increases.
Instead, we propose distilling a lightweight matcher from a VLM which computes correspondences for all entities in a map.
We use this output to form an observation likelihood which is fused over time with a Bayes filter to create a time series of pose estimates.
To support further investigation into generalizable cross-view methods that leverage semantic information, we release a dataset of extracted semantics and evaluation trajectories spanning eleven environments, including panoramas we collected in a snowstorm and at night in Boston. 
We demonstrate our method, trained on a single city's fair-weather data, generalizes across location, lighting, weather, and other challenges. 
Code and datasets are available at \url{https://efahnestock.github.io/loci/}.
\end{keyidea}
\end{abstract}


\section{Introduction}
\begin{keyidea}{}
Robots leverage prior information in the form of maps to enhance environmental awareness and improve navigation performance.
In order for these maps to be useful, the robot must localize itself within the maps.
Maps built from data such as overhead imagery or metric-semantic maps (MSMs), like OpenStreetMap (OSM) \cite{OpenStreetMap}, often give rise to a representation gap: it is difficult to associate online robot sensor data with the contents of the map, complicating localization.
For example, the name of a street in a MSM or how the street appears in overhead imagery can differ substantially in format or perspective from how the street appears in robot sensor data (e.g., pixels in a ground-level panorama).
\end{keyidea}

\begin{keyidea}{Planning with prior data requires localizing against the prior data first; GPS is unreliable}
If the maps are georeferenced, then global navigation satellite systems (GNSS) can sidestep the representation gap by directly providing the robot's location in a global frame.
However, GNSS can be unreliable without direct line of sight to the satellites (e.g., urban canyons) and are susceptible to adversarial jamming.
In this paper, we assume that such absolute positioning sensors are not a viable option.
Without GNSS or a strong prior on the robot's location, we are forced to resolve the representation gap, which can be nontrivial.
Additionally, to be broadly useful, we are interested in approaches that scale to large areas and generalize to new regions without retraining.
\end{keyidea}

\begin{figure}
    \centering
    \includegraphics[width=\linewidth]{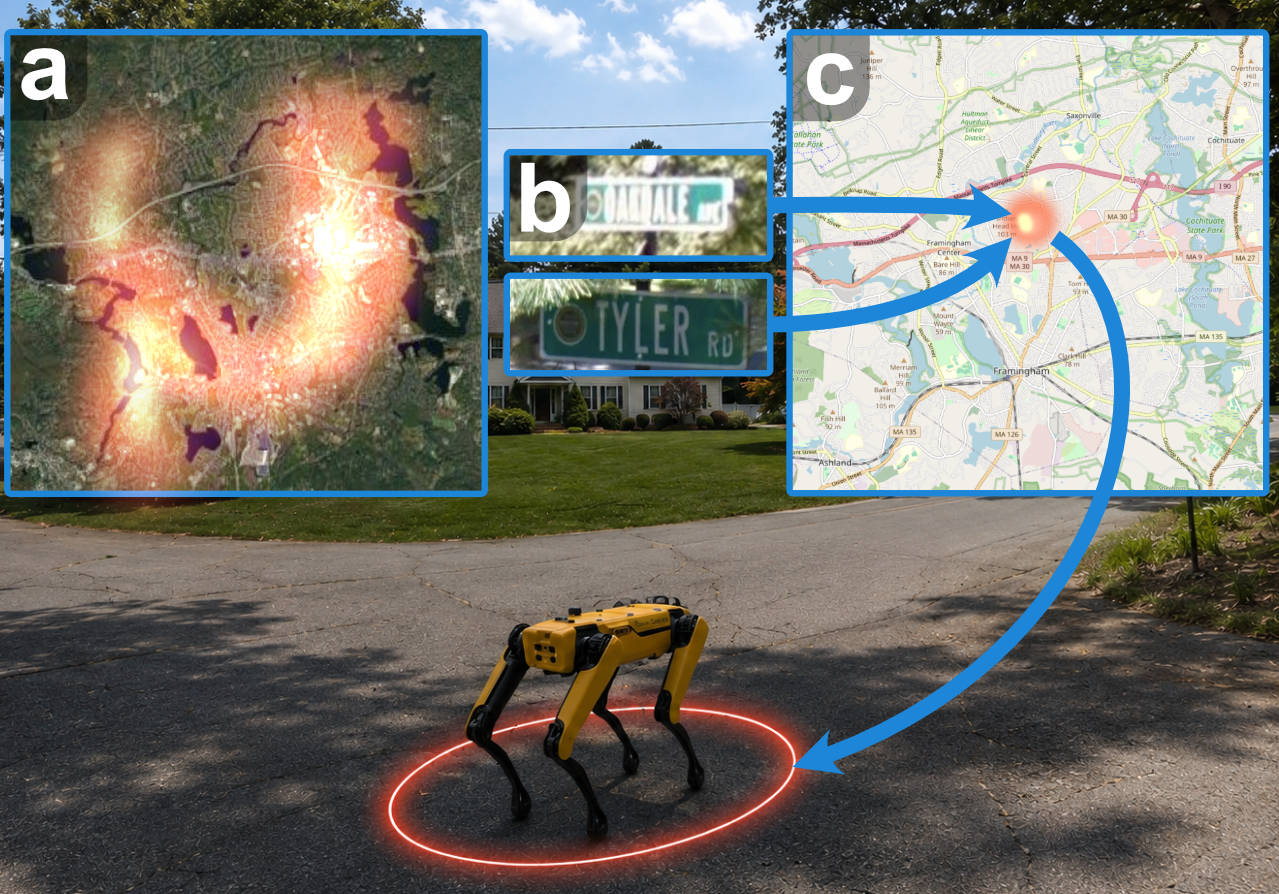}
    \caption{To leverage generally available overhead data, robots need to localize themselves with respect to it. (a) Self-similarity or domain shift can make satellite imagery alone difficult to localize against. (b) Rich semantic features like street names, parks, etc., are often more discriminative and robust to visual change than image-only information. (c) We extract these features and leverage them to more rapidly localize the robot within prior overhead data.}
    \label{fig:intro-figure}
\end{figure}

\begin{keyidea}{Existing Cross-view geo-localization approaches via siamese embeddings; struggles with self-similar areas and generalization }
Cross-view localization \cite{durgam2024cross} directly tackles the representation gap between robot observations and overhead imagery.
Typically, ground and overhead image encoders are trained with a contrastive loss to minimize the distance between embeddings of a ground level observation and its overhead image patch.
This overhead-to-ground image matching approach has seen success on continent scale retrieval tasks \cite{lindenberger_scaling_2025}, but the approach struggles in situations where landmarks present in ground level observations are not visible in the overhead imagery, such as in places with heavy tree cover.
Furthermore, the image matching performance can degrade significantly when the test region differs from the training region due to domain shift in visual appearances \cite{lindenberger_scaling_2025}.
\end{keyidea}

\begin{keyidea}{MSMs have rich semantics but association is hard; previous methods limit to small number of semantic classes, rasterize, and map match. This discards information and fails to scale to larger regions}
MSMs can mitigate some of the shortcomings of image-based cross-view localization.
These maps annotate landmarks at varying levels of granularity, from coarse categories (e.g., building, road, grass, body of water) to instance-level semantics (e.g., name, street address, number of floors).
As with overhead imagery, it is still nontrivial to associate ground-level observations with the semantic features of these metric-semantic maps.

Existing approaches rasterize the MSM into a semantically segmented overhead image, retaining only the coarse categories such as roads, fields, water, etc., and discarding the instance-level semantics.
Association occurs through a dense cross-correlation with bird's eye view projections of robot observations, such as imagery \cite{sarlin_orienternet_2023}, or semantic point clouds \cite{zhou_seglocnet_2025, guo_satellite_2024}.
Since these approaches only rely on the coarse semantic categories, they tend to generalize well to new test regions \cite{sarlin_orienternet_2023, zhou_seglocnet_2025, guo_satellite_2024}.
However, these approaches are only practical for neighborhood-scale environments, or when a strong prior pose belief is available, as the dense cross-correlation scales poorly with the size of the region of interest.
\end{keyidea}

\begin{keyidea}{If we want to use rich semantic information, we need to extract rich semantic information. We also need to match rich semantic information.}
If we wish to extend the use of MSMs to large areas, then we must move beyond coarse semantic classes and leverage the instance-level features contained within these maps.
Knowing that one is at the base of the Eiffel Tower is much more informative than knowing one is next to a building, and yet, current approaches discard precisely this kind of highly specific semantic information. 
The key question is how to bridge the representation gap between a robot's raw sensory observations and the annotations contained within MSMs.
Closing this gap will require the specification of which representations are extracted from the robot observations and the metric-semantic map, and how the representations are compared.
\end{keyidea}

\begin{keyidea}{Our proposed method solves these challenges by: leveraging VLMs to extract semantic information, build a training set for a lightweight classifier to match on semantic information, throw it into a bayes filter}
We address the problem of localization in a Bayesian filtering framework.
In this context, our approach, which we call \textbf{L}andmark-\textbf{O}riented \textbf{C}ross-view \textbf{I}nference (\name{}), is used to provide an observation likelihood given a MSM by bridging the representation gap.
We model the observation likelihood as a product of experts, allowing us to combine existing cross-view localization approaches that provide a likelihood based on overhead imagery, with our model using MSMs to compensate for limitations in the image-based techniques (as illustrated in Figure \ref{fig:intro-figure}).
\name{} leverages vision language models (VLMs), having been trained on images and language, to identify possibly mapped landmarks and produce descriptions in the format of MSM annotations (e.g., \texttt{\{building:tower, name:Eiffel Tower\}}).
VLMs are also effective at determining if two sets of annotations refer to the same landmark \cite{peeters2025entity}.
However, VLMs are not efficient at performing the hundreds of thousands of comparisons that are required for every observed landmark for real-time localization in city-scale environments.
Therefore, \name{} uses a lightweight classifier distilled from a dataset of comparisons performed by a VLM.
We demonstrate in areas as large as 628 $\textrm{km}^2$ that integrating the rich semantics into a localization pipeline can accelerate the filter convergence to the true location in challenging scenarios, including conditions common in field deployments where appearance-based matching is most strained: night and snowstorm operation, heavy rain, fog, post-hurricane environment change, and landmark-sparse rural regions.

Our contributions are as follows:
\begin{itemize}
    \item A method for bridging the representation gap between robot observations and metric-semantic maps like OSM. In particular, we show that a model that aligns panorama image annotations structurally with MSM annotations outperforms unstructured joint panorama/annotation embeddings.
    \item A cross-view dataset targeting conditions relevant to field deployments, including trajectories collected with our own panoramic rig (Figure \ref{fig:collection-rig}), along with VLM-extracted landmarks and labeled correspondence pairs for training and evaluating new semantic cross-view localization approaches.
    \item Open-source tooling to augment existing robotics datasets with OSM landmark annotations and to reproduce all presented results.
\end{itemize}
\end{keyidea}

\section{Problem Formulation}

We formulate cross-view geolocalization as a Bayesian filtering problem: given prior overhead information and a sequence of robot actions and observations, we produce a belief of the robot's current location in the overhead frame.
In this section, we review the Bayesian belief update process and then describe how the prior information and models of the robot and its observations fit within this framework.

\subsection{Bayesian Update of Belief}
\begin{keyidea}{Standard Bayes filter: predict with motion model, use observation likelihoods to incorporate measurements}
A Bayes filter~\cite{thrun_probabilistic_2005} is often used to fuse sequences of actions $u_{0:t-1} \in U^t$ and observations $o_{1:t} \in O^t$ into a belief ${b_t = p(x_t \mid u_{0:t-1}, o_{1:t}, \Prior)} \in \mathcal{B}$ over $x_t \in X$ where prior information $\Prior$ is additional information that is available before the robot is deployed, $X$ is the robot's state space, and $\mathcal{B}$ is the space of beliefs over $X$.
When an observation $o_{t+1}$ is made, the belief can be updated using an observation model $p(o_{t+1} | x_{t+1}, \Prior)$:

\begin{equation}
   b_{t+1} = \eta p(o_{t+1} \mid x_{t+1}, \Prior) p(x_{t+1} \mid u_{0:t}, o_{1:t}, \Prior)
\end{equation}

\noindent where $\eta$ is a normalization constant.
For general distributions, these update equations can be unwieldy.
However, for particular choices of belief, transition dynamics, and observation models, these equations become tractable.

\end{keyidea}

\subsection{Prior Information}
\begin{keyidea}{Prior information can come in many forms (overhead imagery, floor plans, topological maps, topographic maps, etc.) We are primarily concerned with overhead RGB imagery and semantic information coming from open street map.}

Sources of prior information, such as overhead imagery, MSMs, and topographic maps, are rich but partial views of the physical environment, each capturing complementary aspects.
The maps can all be placed into a common world frame, allowing a robot to reason jointly about its pose with respect to the different sources.
In this work, we focus on overhead imagery and MSMs, which we refer to collectively as $\Prior = (\I, \M)$.

Overhead imagery $\I \in \mathbb{R}^{H \times W \times 3}$ is a georeferenced image where each pixel corresponds to a known location in the world and its value indicates the color observed by a camera at that location.
An image patch centered at location $p \in \R^{2}$ is denoted as $\mathcal{I}_p \in \mathbb{R}^{h \times w \times 3}$ where $h \leq H$ and $w \leq W$.

A metric-semantic map $\mathcal{M} = \{\ell_i^\M\}_{i=1}^N$ is a set of landmarks where each landmark $\ell^\M = (g_{\ell}, a_{\ell})$ has geometry $g_{\ell}$ in a common frame and annotations $a_{\ell}$.
The geometry $g_{\ell}$ is a geometric primitive in $\R^2$ (e.g., point, polyline, polygon, etc.).
The annotations $a_{\ell} = \{(k_j, v_j)\}_{j=1}^{N_{\ell}}$ are a set of key-value pairs describing the properties of the landmark (e.g. $a_{\ell}$ = \texttt{\{amenity:restaurant, cuisine:Vietnamese\}}).\footnote{Examples of MSMs include GIS maps, HD maps, object-level semantic SLAM maps, scene graphs, and OpenStreetMap. Maps that don't meet this definition include topological maps (no geometric information), point clouds, voxel grids, and rasterized maps (no discrete landmarks).}
\end{keyidea}

\subsection{Robot Model}
\label{sec:robot_model}
\begin{keyidea}{Robot state and transition model}
We take the set of robot poses $X$ and actions $U$ to be subsets of $\R^2$.
The heading is omitted from the state, as justified in Section \ref{sec:obs_model}.
A robot takes an action $u_t \in U$ and the stochastic transition model $p(x_{t+1} | x_t, u_t)$ is given by $ x_{t+1} = f(x_t, u_t, v_t)$ where $x_t$ and $u_t$ are the current state and applied action and $v_t \sim \mathcal{N}(0, I)$ is a noise parameter.
\end{keyidea}

\subsection{Observation Model}
\label{sec:obs_model}
\begin{keyidea}{A robot has a panoramic camera, we assume that the observations are made aligned with the world frame.}
We assume that the robot is equipped with a panoramic camera, which produces observations $o \in O \subset \R^{H_c \times W_c \times 3}$ according to the distribution $p(o|x, \Prior)$.
$H_c$ and $W_c$ are the height and width of the image. 
We further assume that the observations can be axis aligned to the world frame (e.g., via a magnetometer). As a result, the robot heading does not affect the observation likelihood and is omitted from the robot state.
\end{keyidea}

\begin{keyidea}{While can get samples from our observation model, to perform Bayesian filtering, we at a minimum need to compute the unnormalized density}

During a deployment, the robot takes observations $o$, and we assume these are generated according to the observation model $p(o | x, \Prior)$.
However, in order to perform Bayesian filtering, we need to be able to evaluate the density of the observation model at the observed panorama, up to a normalization constant.
The key challenge addressed by this paper is how to specify this observation model.
\end{keyidea}

\section{Approach}
\label{sec:approach}
\begin{figure*}[t]
    \centering
    \input{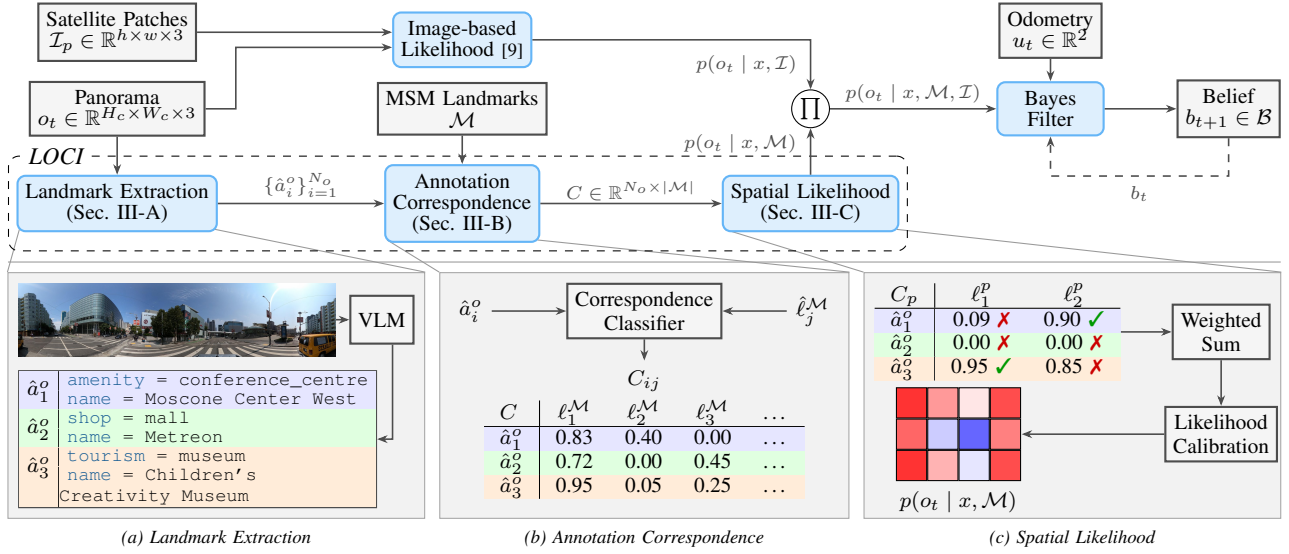}
    \caption{(Top) System diagram of the proposed approach. An image-based observation likelihood $p(o | x, \I)$ from WAG \cite{downes_city-wide_2022} is fused with \name{}'s MSM-based likelihood to form the combined likelihood $p(o | x, \I, \M)$. A Bayes filter incorporates this likelihood with odometry $u_t$ to update the pose belief $b_{t+1}$. (a) \name{} processes an observation $o$ with a VLM to extract landmark annotations $\{\hat{a}^o_i\}_{i=1}^{N_o}$. (b) A correspondence classifier then matches each annotation $\hat{a}^o_i$ against MSM landmarks $\M$. (c) A spatial likelihood module aggregates the correspondence scores into per-patch observation probabilities $p(o_t | x, \M)$.}
    \label{fig:late_fusion_system}
\end{figure*}

\begin{keyidea}{\name{} models the observation likelihood as a product of experts}

Rather than specify $p(o \mid x, \Prior)$ directly, we approximate the observation likelihood as a product of experts \cite{hinton_training_2002}: 
\begin{equation}
p(o \mid x, \Prior) =  p(o \mid x, \I, \M) \propto p(o\mid x, \I) p(o\mid x, \M),
\end{equation}
where each term relates the observation to a source of prior information.
This structure is exact when $\I$ and $\M$ are conditionally independent given the location and panorama. 
We adopt this structure because it lets us build each expert independently and because the two experts contribute complementary signal\footnote{A sign with the name of a street is apparent in the MSM but not the overhead imagery. Conversely, an unmarked trail can be visible from the overhead imagery but not the MSM.}.
However, in practice, they are correlated views of the same scene\footnote{A church that is apparent in both the overhead imagery and the MSM would yield an overconfident observation likelihood.}, so the product over-counts shared evidence, yielding overconfident estimates.
Despite this shortcoming, we find that this structured approach outperforms an approach that learns the jointly conditioned likelihood (see Section \ref{sec:evaluation}). 
As illustrated in Figure \ref{fig:late_fusion_system}, \name{} is implemented as a three-stage pipeline.
First, the panorama is processed by a VLM, yielding a set of extracted landmarks.
Next, similarity scores between each extracted landmark and each landmark in the MSM are computed.
Finally, the similarity scores are used to compute the MSM-based likelihood.
This MSM-based likelihood is then combined with an overhead-imagery-based observation likelihood. The overhead-imagery-based likelihood is provided by Wide Area Geolocalization (WAG) \cite{downes_city-wide_2022}.
Each stage is now described in turn, starting with landmark extraction from panoramas.
\end{keyidea}

\subsection{Landmark Extraction From Panoramas}
\label{sec:landmark_extraction}

\begin{keyidea}{What representation do we extract}
Consider a physical landmark which is observed by the robot and is also described by the MSM. 
Let $o_\ell$ denote the landmark's visual appearance and $\ell^\M$ its corresponding entry in the MSM.
The detected object $o_\ell$ and mapped landmark $\ell^\M$ are fundamentally different: one is a region of an image and the other is a pair of the landmark geometry and semantic annotations.
To assess whether they refer to the same landmark, we must first map them to a common representation that admits comparison.

One could imagine a system that encodes $o_{\ell}$ and $\ell^\M$ into a joint embedding space, for example by using readily available sentence embedding models to embed natural language descriptions of $o_\ell$ and $\ell^\M$.
However, this approach collapses the landmark's attributes such as name and category into a single descriptor.
As a result, ``a McDonald's fast food restaurant'', ``a Wendy's fast food restaurant'', and ``Old MacDonald's Farm'' could be deemed similar, even though in a localization context, these are all distinct landmarks.

In contrast, the MSM schema keeps these different concepts separate.
We are able to reason that although an \texttt{amenity} tag is shared by \texttt{\{name:McDonald's,\allowbreak amenity:restaurant\}} and \texttt{\{name:Wendy's,\allowbreak amenity:restaurant\}}, the names identify them as separate landmarks.
Similarly, although \texttt{\{name:McDonald's,\allowbreak amenity:restaurant\}} and \texttt{\{name:Old MacDonald's,\allowbreak landuse:farmland\}} have similar names, the \texttt{amenity} and \texttt{landuse} tags clearly distinguish them.
Therefore, we choose to have a VLM produce annotations aligned with the structure of those contained in MSMs.
\end{keyidea}

\begin{keyidea}{The VLM extracts annotations given a panorama and a text prompt}

\begin{table}
\centering
\caption{Structure of the VLM prompt for landmark extraction from panoramic observation.}
\label{tab:prompt-structure}
\begin{tabular}{p{3cm} p{5cm}}
\toprule
\textbf{Prompt Component} & \textbf{Description} \\
\midrule
Panorama Images & Pinhole projections of the panorama. \\
\addlinespace
Task Instructions & General task description, recommended workflow, and desired output information. \\
\addlinespace
Landmark Selection & Landmark specific criteria for selection. \\
\addlinespace
MSM Ontology & MSM Tagging background and examples. \\
\addlinespace
Exclusion Criteria & Landmark-independent filtering rules. \\
\addlinespace
Output Schema & JSON schema describing the output format. \\
\bottomrule
\end{tabular}
\end{table}

Concretely, given a robot observation $o$, we want the VLM to identify relevant landmarks and produce $\hat{a}^o = \{\hat{a}^o_i\}_{i=1}^{N_o}$, where $\hat{a}^o$ is a set of annotations describing a landmark and $N_o$ is the number of landmarks identified in $o$.
The VLM receives two kinds of inputs: a set of pinhole projections from the panorama, and a structured text prompt outlined in Table \ref{tab:prompt-structure}. 
We provide four square pinhole projections each with $90^\circ$ field of view from the panoramic observations as the distortion introduced by the equirectangular projection is likely to be out of distribution for the VLM.
The VLM prompt provides task instructions, MSM ontology examples, landmark selection and exclusion criteria, and a target output schema.
The instructions ask the VLM to extract useful landmarks (e.g., street names, branded locations, architecturally unique buildings) while ignoring less informative ones (e.g., traffic signals, trees, buildings described only by appearance).
We also instruct the VLM to ignore distant landmarks, transient landmarks, and illegible signs.
\end{keyidea}

\subsection{Correspondence of Annotations to Prior Information}
\label{sec:landmark_correspondence}
\begin{keyidea}{OSM maps are unreliable: missing tags, missing landmarks, outdated landmarks}
Having extracted a set of annotations from the panorama, we now need to match them against entries in the MSM. For each annotation $\hat{a}_i^o$ extracted from the panorama and each landmark $\ell^{\M}_j$ in the MSM, we would like to compute a correspondence score $C_{ij}$ that is maximal when $\hat{a}_i^o$ and $\ell^{\M}_j$ refer to the same physical landmark.
When approaching this problem, it is important to consider what goes into an MSM.
These maps are often crowd-sourced, meaning which landmarks are annotated and how they are annotated is up to the individual contributor.
Annotation conventions have developed over time, but any given landmark may deviate from these established conventions.
As a result, different instances of the same kind of landmark may be annotated differently.
For example, an apartment building might be richly annotated with its date of construction, number of floors and architecture style, and another apartment building might have no information other than that the building exists.
Given the variation in annotations, it is difficult to write down criteria for when two sets of annotations describe the same physical landmark.

\end{keyidea}
\begin{keyidea}{VLMs are okay at matching, but too expensive to run -> Create a dataset}
VLMs are proficient at judging whether two descriptions could refer to the same entity~\cite{peeters2025entity}, a capability we leverage to match landmark annotations.
However, the latency and expense of querying a VLM to perform the hundreds of thousands of comparisons required for each observed landmark is untenable for a robot deployment (see runtime reported in Section \ref{sec:evaluation_details}).
Therefore, we use the VLM as a teacher to generate training examples of positive and negative matches offline.
We then distill a lightweight classifier $p(c_{ij} \mid \hat{a}_i^o, \ell_j^\M)$ suitable for online inference, where $c_{ij}$ is a binary random variable indicating correspondence between an annotation $\hat{a}_i^o$ and landmark $\ell_j^{\M}$.
\end{keyidea}

\begin{keyidea}{A VLM teacher mines positive matches and hard negatives}

\begin{table}
\centering
\caption{Structure of the VLM prompt for annotation correspondence labeling.}
\label{tab:correspondence-prompt-structure}
\begin{tabular}{p{3cm} p{5cm}}
\toprule
\textbf{Section} & \textbf{Description} \\
\midrule
Task Instructions & Definition of the matching task on annotations, with the caveat that some landmarks have no counterpart. \\
\addlinespace
Hard-Negative Criteria & Explicit rules for what counts as a hard negative and which tag-level disagreements to ignore. \\
\addlinespace
Set 1 (panorama) & Numbered list of \texttt{key=value} bundles for the panorama-extracted annotations. \\
\addlinespace
Set 2 (map) & Numbered list of \texttt{key=value} bundles for nearby map landmarks. \\
\addlinespace
Output Schema & JSON schema requiring, for each Set~1 entry, the matching Set~2 indices and 0--2 negatives labeled \emph{hard} or \emph{easy}. \\
\bottomrule
\end{tabular}
\end{table}

To assemble a training set for this classifier, we pair the set of landmarks detected by the VLM in each panorama at a known location in the training city with the set of MSM landmarks near that location.
The teacher receives both sets in their native \texttt{key=value} form, using the same structured representation that the distilled classifier will consume.
The prompt, summarized in Table~\ref{tab:correspondence-prompt-structure}, supplies a task description, the two numbered tag-bundle lists Set~1 (panorama landmarks) and Set~2 (map landmarks), and a JSON output schema.

For each panorama, we ask the teacher to produce, for each Set~1 entry, the indices of any Set~2 landmarks that refer to the same physical object and up to two explicit negatives drawn from Set~2.
Negatives are labeled \emph{easy} when the two annotations describe obviously different kinds of objects (e.g., a restaurant versus a street) and \emph{hard} when they share a general category but unambiguously refer to different instances such as distinct branded names or vastly different number of floors.
The prompt explicitly tells the teacher to \emph{not} mine hard negatives from disagreements that the annotation extractor is known to be unreliable about, such as small numeric differences in \texttt{building:levels}, equivalent tags of different specificity (\texttt{building=apartments} versus \texttt{building=yes}), or one name being a substring of another.
This prompting strategy shapes the negative set toward the failure modes that the lightweight classifier actually needs to learn to distinguish, rather than penalizing it for noise that the extractor introduces.
The resulting $(\hat{a}_i^o, \ell^{\M}_j, c_{ij})$ triples --- positives from the VLM's match list, hard and easy negatives from its negative list --- are then used to train the lightweight classifier.
\end{keyidea}
\begin{keyidea}{Lightweight model structure}
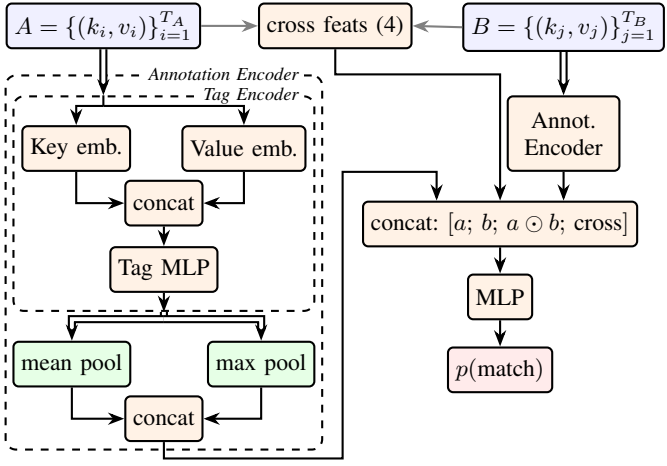
\begin{figure}[t]
    \centering
    \resizebox{\linewidth}{!}{\begin{tikzpicture}[
  font=\small,
  >={Stealth[length=2.2mm]},
  every path/.style={thick},
  data/.style   ={draw,rounded corners=2pt,inner sep=3pt,
                  minimum height=6mm,fill=blue!6},
  op/.style     ={draw,rounded corners=2pt,inner sep=3pt,
                  minimum height=6mm,fill=orange!10},
  pool/.style   ={draw,rounded corners=2pt,inner sep=3pt,
                  minimum height=6mm,fill=green!10},
  final/.style  ={draw,rounded corners=2pt,inner sep=3pt,
                  minimum height=6mm,fill=red!8},
  plate/.style  ={draw,dashed,rounded corners=4pt,inner sep=0pt},
  dim/.style    ={font=\scriptsize\itshape,inner sep=1pt},
  edgelbl/.style={font=\scriptsize\itshape,inner sep=1pt,fill=white},
]

\newif\ifdebug \debugfalse

\coordinate (north-west) at (0,0);
\coordinate (north-east) at (\columnwidth,0);
\coordinate (north) at (\columnwidth/2,0);

\node[data, anchor=north west] (A) at (north-west)   {$A=\{(k_i,v_i)\}_{i=1}^{T_A}$};
\node[data, anchor=north east] (B) at (north-east) {$B=\{(k_j,v_j)\}_{j=1}^{T_B}$};
\node[op]   (cross) at (north |- A) {cross feats (4)};
\draw[->,gray] (A.east) -- (cross.west);
\draw[->,gray] (B.west) -- (cross.east);

\def\platesep{3pt}
\def\topplatesep{8pt}
\coordinate (annot-north-west) at ($(A.south west) + (0, -10pt)$);
\coordinate (annot-south-east) at ($(annot-north-west) + (\columnwidth/2 - 5pt, -143pt)$);
\coordinate (tag-north-west) at ($(annot-north-west) + (\platesep, -\topplatesep)$);
\coordinate (tag-south-east) at ($(annot-south-east |- tag-north-west) + (-\platesep, -82pt)$);

\node[plate, fit=(annot-north-west)(annot-south-east)] (annot-detail-plate) {};
\node[font=\scriptsize\itshape,fill=white,inner sep=2pt,anchor=east] (annot-encoder-name)
      at ($(annot-detail-plate.north east) + (-7pt, 0)$) {Annotation Encoder};

\node[plate, fit=(tag-north-west)(tag-south-east)] (tag-detail-plate) {};
\node[font=\scriptsize\itshape,fill=white,inner sep=2pt,anchor=east]
      at ($( annot-encoder-name.east |- tag-detail-plate.north east)$) {Tag Encoder};

\def\toptagplatesep{11pt}
\node[op, anchor=north west] (keyA) at ($(tag-north-west) + (\platesep, -\toptagplatesep)$) {Key emb.};
\node[op, anchor=north east] (txtA) at ($(tag-south-east |- keyA.north) + (-\platesep, 0)$) {Value emb.};

\draw[->,double,double distance=1.5pt] (A.south) -- (A.south |- tag-detail-plate.north);

\coordinate (arrow-split) at ($(A |- keyA.north) + (0, \toptagplatesep*2/3)$);
\draw[->] (arrow-split) -- (keyA |- arrow-split) -- (keyA);
\draw[->] (arrow-split) -- (txtA |- arrow-split) -- (txtA);
\draw[-] (A |- tag-detail-plate.north) -- (arrow-split);

\node[op, anchor=north] (concatTag) at ($(tag-detail-plate |- keyA.south) + (0, -\platesep)$) {concat};
\draw[->] (keyA) -- (keyA |- concatTag) -- (concatTag);
\draw[->] (txtA) -- (txtA |- concatTag) -- (concatTag);

\node[op, anchor=north] (mlpA) at ($(concatTag.south) + (0, -2.5*\platesep)$) {Tag MLP};
\draw[->] (concatTag) -- (mlpA);

\draw[->] (mlpA) -- (tag-detail-plate.south);

\node[pool, anchor=north west] (meanA) at ($(annot-detail-plate.west |- tag-detail-plate.south) + (\platesep, -\toptagplatesep)$) {mean pool};
\node[pool, anchor=north east] (maxA)  at ($(annot-detail-plate.east |- tag-detail-plate.south) + (-\platesep, -\toptagplatesep)$) {max pool};

\coordinate (pool-split) at ($(tag-detail-plate.south) + (0, -1/4*\toptagplatesep)$);
\draw[-, double, double distance=1.5pt] (tag-detail-plate.south) -- (pool-split);
\draw[->, double, double distance=1.5pt] (pool-split) -- (meanA |- pool-split) -- (meanA);
\draw[->, double, double distance=1.5pt] (pool-split) -- (maxA |- pool-split) -- (maxA);

\node[op, anchor=north] (concatA) at ($(annot-detail-plate |- meanA.south) + (0, -\platesep)$) {concat};
\draw[->] (meanA.south) |- (concatA.west);
\draw[->] (maxA.south)  |- (concatA.east);

\node[op,minimum width=1.5cm,minimum height=1.0cm, anchor=north,
      align=center] (encB) at (B.south |- tag-detail-plate.north)
      {Annot.\\Encoder};
\draw[->,double,double distance=1.5pt] (B.south) -- (encB.north);

\coordinate (combine-x) at ($(cross.south)!1/2!(B.south east)$);
\coordinate (combine-north) at ($(combine-x |- encB.south) + (0, -\toptagplatesep)$);
\node[op, anchor=north] (combine) at (combine-north)
      {concat: $[a;\,b;\,a\odot b;\,\text{cross}]$};

\coordinate (cross-bend) at (cross.south |- annot-detail-plate.north);
\draw[->] (encB) -- (encB |- combine.north);
\draw[->] (cross) -- (cross-bend) -| (combine);

\coordinate (concatA-bottom-bend) at ($(annot-detail-plate.south) + (0, -\platesep)$);
\coordinate (between-annot-combine) at ($(annot-detail-plate.east)!1/2!(combine.west)$);
\coordinate (between-encB-combine) at (encB.south);
\coordinate (concatA-top-bend) at (between-annot-combine |- between-encB-combine);

\coordinate (encB-combine-meet) at (encB |- combine.north);
\coordinate (encA-combine-meet) at ($(combine.north)!-1!(encB-combine-meet)$);
\draw[->] (concatA) -- (concatA-bottom-bend) -| (concatA-top-bend) -| (encA-combine-meet);
\node[op, anchor=north] (mlp) at ($(combine.south) + (0, -10pt)$) {MLP};
\node[final, anchor=north] (sigout) at ($(mlp.south) + (0, -10pt)$) {$p(\text{match})$};
\draw[->] (combine.south) -- (mlp.north);
\draw[->] (mlp.south) -- (sigout.north);

\ifdebug
      \draw[red,thin] (0, 0) -- (0, -200pt);
      \draw[red,thin] (\columnwidth, 0) -- (\columnwidth, -200pt);
      \coordinate (concat-south) at ($(concatA.south) + (0, -\platesep)$);
      \draw[red,thin] (north-west |- concat-south) -- (north-east |- concat-south);

\fi

\end{tikzpicture}}
    \caption{Network architecture for the correspondence classifier. This classifier takes in two sets of annotations ($A, B$) and predicts if the two sets of annotations could refer to the same physical landmark. An annotation encoder embeds then pools across each set of annotations to create a descriptor. The descriptors for the two sets, alongside their Hadamard product and 4 cross features are fed to a final MLP to produce the probability the landmarks described by $A$ and $B$ could match.}
    \label{fig:landmark_correspondence_model}
\end{figure}

The lightweight classifier consumes two sets of annotations and predicts if the annotations could describe the same physical landmark. 
The network architecture for the correspondence classification model is illustrated in Figure \ref{fig:landmark_correspondence_model}.
Each key-value pair is first encoded by separately encoding the key and the value.
To encode the key, we map each of a set of 108 predetermined keys to a unique learned 32-dimensional embedding. 
To encode the value, we use Google's \texttt{text-embedding-005} model to produce a 768-dimensional embedding that is projected down to a 128-dimensional vector.
These two embeddings are concatenated and then projected down to a 64-dimensional embedding with an MLP.
After performing this concatenation and projection for each key-value pair in an annotation set, the embeddings are combined via mean and max pooling.
These pooled vectors are concatenated to form an annotation embedding, one for each of the two annotation sets.
In addition, the two sets of annotations are used to compute a set of cross features.
These cross features include the mean value cosine similarity among key-value pairs that share a key, the max value cosine similarity among the same pairs, the cosine similarity if both annotations contain a \texttt{name} key, and finally, a boolean flag denoting if both sets contain a compatible \texttt{addr:housenumber} field.
Two house number fields are compatible if they are equal, or if either/both of the values represents a range, if the ranges overlap.

The embeddings for the two sets of annotations, their Hadamard product, and the cross features are concatenated to form the input for an MLP which produces the confidence that both annotations could describe the same landmark.

With this correspondence classifier, we can now produce a score matrix $C \in \mathbb{R}^{N_o\times |\M|}$ where the rows correspond to the annotations extracted from the panorama and the columns correspond to mapped landmarks.
Each entry of the matrix is the confidence that the relevant row and column describe the same physical landmark.
We now show how this score matrix can be used to produce the observation likelihood $p(o|x, \M)$.
\end{keyidea}

\subsection{Observation Likelihood from Matches}
\label{sec:landmark_observation_likelihood}

\begin{keyidea}{Given the probability of the extracted landmarks matching a mapped landmark, we want to create an observation likelihood given location and the map}

In the last stage of the pipeline, given a score matrix, we compute the observation likelihood $p(o \mid x, \M)$. 
The most faithful model of the observation likelihood would include a location conditioned occlusion model, a noise model for VLM extraction, and a model of map incompleteness.
Specifying each of these components would require extensive modeling effort.
In addition, evaluating such a model would likely be prohibitively expensive as the area of interest grows larger. 
\end{keyidea}

\begin{keyidea}{A tractable approximation}
As a tractable approximation, we discretize the world into patches of fixed size, and visibility is approximated as constant for features within the area of the observer's patch and zero elsewhere.
We now show how this approximation is used to compute $p(o \mid x, \M)$.
To compute a panorama-patch score between panorama $o$ and overhead image patch $\I_p$ centered at $p$, we start with the set of annotations $\hat{a}^o = \{\hat{a}^o_i\}_{i=1}^{N_o}$ extracted from the panorama, and the set of MSM landmarks $\M_p$ located within patch $\I_p$. We then collect the relevant correspondence scores $C_p \in \mathbb{R}^{N_o \times |\M_p|}$ described in Section \ref{sec:landmark_correspondence}, where the rows correspond to annotations $\hat{a}^o$ and the columns correspond to the mapped landmarks that overlap with $\I_p$.
We then use the Hungarian algorithm \cite{kuhn_hungarian_1955} with these correspondence scores to determine the most compatible assignment of extracted annotations to mapped landmarks.
To avoid low confidence matches, the matrix is augmented with dummy rows and columns such that each annotation can be assigned to a dummy landmark at some fixed cost \cite{sarlin2020superglue}.
Finally, we perform a weighted sum of the correspondence scores for the pairs selected by the Hungarian algorithm.
We weight each correspondence score by $1/\log_2(1+\max(1, N_{\text{matches},i}))$ where $N_{\text{matches},i}$ is the number of MSM landmarks whose correspondence score with the annotations $\hat{a}^o_i$ exceeds the Hungarian matching threshold. This weighting scheme makes distinctive landmarks more impactful on the final matching score. We now have a score $s_{o, p} \in \mathbb{R}^+$ for every overhead patch $\I_p$.
\end{keyidea}

\begin{keyidea}{Patch Scores to Observation Likelihoods}
Inspired by Downes \textit{et al.}~\cite{downes_city-wide_2022}, we use a Gaussian likelihood to model the distribution of scores. To compute the observation likelihood $p(o \mid x, \M)$, we compute:
\begin{equation} \label{eq:obs_liklihood_landmark}
p(o\mid x_i, \M) \propto \exp{\frac{-1}{\sigma_l^2} \left(1 - \frac{s_{o,i}}{\max_j s_{o, j}}\right)^2}.
\end{equation}

Now that we have a landmark-based observation likelihood, as illustrated in Figure \ref{fig:late_fusion_system}, it can be fused with the overhead imagery-based likelihood to accelerate filter convergence.
Next, we describe the datasets we construct to train and evaluate \name{}.
\end{keyidea}

\section{\name{} Dataset} \label{sec:dataset}

\begin{keyidea}{Motivate generally + release statement, enumerate the three artifacts}
The first step to training and evaluating semantic landmark-based cross-view localization approaches is to gather data.
The datasets for cross-view localization have become more numerous in recent years %
but often consist solely of imagery and lack semantic landmarks.
Datasets that contain semantic landmarks both focus on fair-weather panoramas and leverage privileged information during generation (e.g., providing the nearby OSM landmarks while extracting semantic descriptions from a panorama) \cite{ye_where_2025}, making extracted scene descriptions unrealistic for new environments. 

To support future work in landmark-based cross-view localization in a range of environments, we release three artifacts: a curated set of GPS-stamped panoramas with associated OSM landmarks and overhead satellite images, a set of VLM extracted annotations from these panoramas, and finally a dataset of annotation correspondences used to train and evaluate our correspondence model.
All of the processes, code, and data can be found at the Github repository linked in the abstract.
In the following three sections we describe each of these three artifacts.
General dataset statistics are detailed in Table \ref{tab:dataset-stats}.
\end{keyidea}

\begin{table*}[t]
  \centering
  \caption{Dataset statistics. \textsuperscript{\textdagger} indicates data collected by our camera rig.}
  \label{tab:dataset-stats}
  \small
  \begin{tabular}{llrrrrc}
  \toprule
  City                                      & Conditions                   & \#Panos / Sat.\ Patches & \makecell{Sat.\ Cov.\\(km$^2$)} & Traj.\ (km) & \makecell{Landmarks\\(Pano / OSM)} & \makecell{Dates (YY-MM) \\(Pano / OSM)} \\
  \midrule
  Chicago (VIGOR)                           & Metro area                   & 25{,}479 / 22{,}308     & 30.1                            & --          & 62{,}594 / 57{,}531                & 2019 / 20-01                            \\
  Seattle (VIGOR)                           & Metro area                   & 23{,}751 / 20{,}776     & 28.0                            & --          & 42{,}124 / 60{,}964                & 2019 / 20-01                            \\
  New York (VIGOR)                          & Metro area                   & 27{,}769 / 23{,}279     & 59.6                            & --          & 95{,}462 / 109{,}548               & 2019 / 20-01                            \\
  San Francisco                             & Metro area                   & 300 / 24{,}255          & 32.8                            & 5.2         & 1{,}434 / 113{,}387                & 21-08 / 22-01                           \\
  Boston Snowy\textsuperscript{\textdagger} & Actively snowing             & 1{,}674 / 44{,}290      & 62.4                            & 12.4        & 6{,}159 / 162{,}558                & 26-01 / 26-01                           \\
  Boston Night\textsuperscript{\textdagger} & Recorded at night            & 1{,}378 / 44{,}290      & 62.4                            & 12.5        & 4{,}028 / 162{,}558                & 26-02 / 26-01                           \\
  Middletown                                & Rainy, suburban              & 263 / 39{,}601          & 51.0                            & 6.9         & 233 / 20{,}256                     & 25-04 / 25-01                           \\
  Framingham                                & Suburban                     & 478 / 40{,}401          & 50.9                            & 5.6         & 209 / 28{,}827                     & 25-08 / 26-01                           \\
  Fort Myers                                & Post-disaster                & 1{,}073 / 471{,}822     & 627.6                           & 11.8        & 1{,}365 / 119{,}990                & 22-10 / 22-01                           \\
  Noordoostpolder                           & Foggy, rural, suburban & 1{,}916 / 492{,}804     & 627.0                           & 13.0        & 2{,}508 / 162{,}263                & 25-01 / 25-01                           \\
  Veluwe                                    & Highway, forest, rural & 3{,}654 / 494{,}209     & 628.4                           & 25.0        & 3{,}607 / 419{,}969                & 25-02 / 25-01                           \\
  \bottomrule
  \end{tabular}
\end{table*}

\subsection{Panoramas and Overhead Data}

\begin{keyidea}{Three pano sources: VIGOR, Mapillary, self-collected}
The base of the dataset is collections of north-aligned geo-located egocentric panoramas $o_t \in \mathbb{R}^{H_c \times W_c \times 3}$. We source these panoramas from the VIGOR dataset \cite{zhu_vigor_2021}, Mapillary (a crowd-sourced street-level imagery platform) \cite{MapillaryPlatform}, and through our own data collection. 
The VIGOR dataset \cite{zhu_vigor_2021} is a foundational dataset in the field of cross-view localization and provides densely sampled panoramas in fair-weather daytime conditions in four cities in the United States (Chicago, Seattle, New York and San Francisco). VIGOR dataset panoramas are north-aligned (where the center of the panorama points north) and are $2048 \times 1024$\,px. Note we do not use the San Francisco VIGOR panoramas in this paper. 
\end{keyidea}

\begin{keyidea}{Diversity: Mapillary + Boston collects probe weather, lighting, disaster}
Generalization to novel conditions and environments is a key focus of our work, so we curate evaluation trajectories of GPS-stamped panoramas in settings beyond fair-weather urban environments. We are further motivated by the resolution of VIGOR panoramas, which is often limiting for landmark extraction. We source these higher-resolution trajectories from Mapillary \cite{MapillaryPlatform} and by recording our own GPS-stamped panoramas with the rig shown in Figure \ref{fig:collection-rig}.

\begin{figure}[t]
    \centering
    \includegraphics[width=\linewidth]{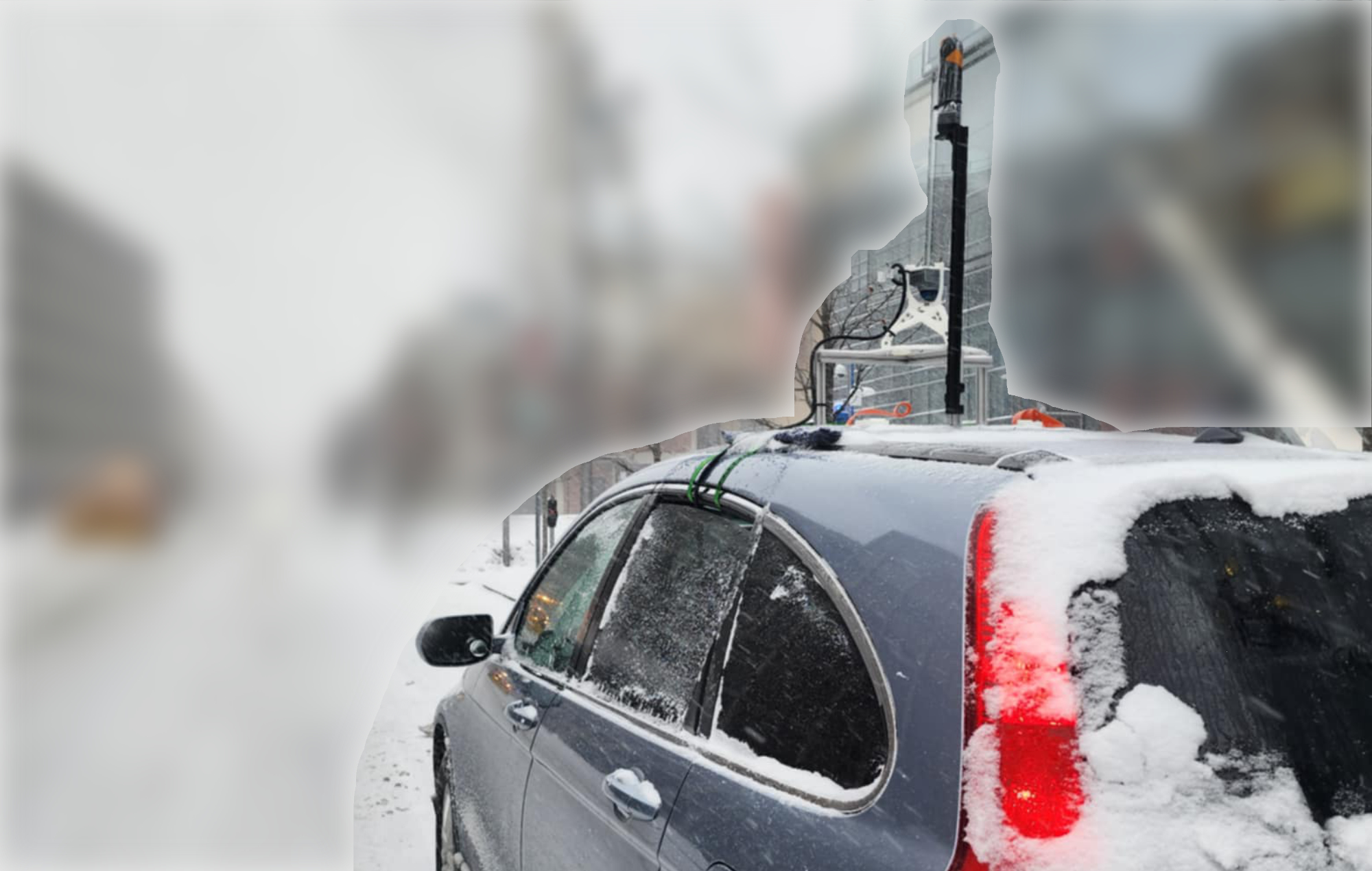}
    \caption{Panorama collection rig used for Boston data collects. A Qoocam 8k Enterprise 360 degree camera mounted on the roof captured panoramas while a u-blox F9R GPS receiver inside the car recorded GPS position. The Livox lidar was only used to later align the camera and GPS data by matching the point of initial movement.}
    \label{fig:collection-rig}
\end{figure}

Specifically, we investigate robustness to weather with a Mapillary trajectory collected during a rain storm in Middletown (CT, USA) and a trajectory we recorded during a snowstorm in Boston (MA, USA).
We investigate impact of lighting with a trajectory we collect at night in Boston, and performance in very rural to more urban environments with Mapillary trajectories from Veluwe (Netherlands), Noordoostpolder (Netherlands), Framingham (MA, USA), and San Francisco (CA, USA) respectively.
We also explore robustness to non-trivial environment change after a natural disaster with Fort Myers (FL, USA), where the Mapillary trajectory was collected shortly after Hurricane Ian passed through the area.
Trajectories vary in length from 5\,km to 25\,km, and generally do not contain large loops with the exception of San Francisco. 
\end{keyidea}

Mapillary provides an estimated compass angle for each of its panoramas, which was used to north-align each image, matching the panoramas in the VIGOR dataset. Resolution varies by Mapillary trajectory with a minimum resolution of $4096 \times 2048$\,px. For the self-collected panoramas, heading was derived from sequential GPS readings and image resolution was $7680 \times 3840$\,px.

\begin{keyidea}{Satellite + OSM sourcing}
Each collection of panoramas (each VIGOR city, along with the Mapillary and recorded trajectories), is then paired with overhead imagery $\I$ covering the area we wish to localize within, and a metric-semantic map $\M$ for that same area.
The VIGOR dataset already provides $640\times640$\,px satellite patches at Web Mercator zoom level 20, which translates to each individual satellite patch being \textasciitilde73 m on each side at latitudes similar to the city of Chicago. Satellite patches overlap by $50$\% with all of their neighboring patches \cite{zhu_vigor_2021}.

For the Mapillary trajectories and our recorded panorama trajectories, overhead regions encompassing large areas (varying from 32\,km$^2$ to 628\,km$^2$) around the trajectories were defined. Satellite images were collected from distributable sources: MassGIS spring 2025 \cite{massgis_2025_imagery} imagery was used for Framingham, Connecticut spring 2023 state-wide orthoimagery was used for Middletown \cite{ct_ortho_2023}, VIGOR's satellite imagery was used for San Francisco, and ESRI World Imagery pinned to Wayback releases \cite{esri_wayback_hurricane_ian} was used for Boston Snowy/Night (capture date 2023-05), Fort Myers (2021-01, before Hurricane Ian), Noordoostpolder (2025-01), and Veluwe (2025-01). Additionally, satellite patches for Noordoostpolder, Veluwe, and Fort Myers were cropped or upsampled to maintain roughly 73\,m a side because of their latitude difference from the Northern US. The same $640\times640$\,px and $50\%$ overlap as the VIGOR overhead satellite patches was maintained.

We source our MSM data from OpenStreetMap for each environment using GeoFabrik's historical OSM backups of each region, allowing the state of the OSM landmarks to coarsely align with when the panorama images were captured. All landmarks with coordinates lying outside of the specified overhead imagery coverage were discarded, resulting in the final landmark statistics shown in Table \ref{tab:dataset-stats}.
\end{keyidea}

\begin{figure}[t]
    \centering
    \includegraphics[width=\linewidth]{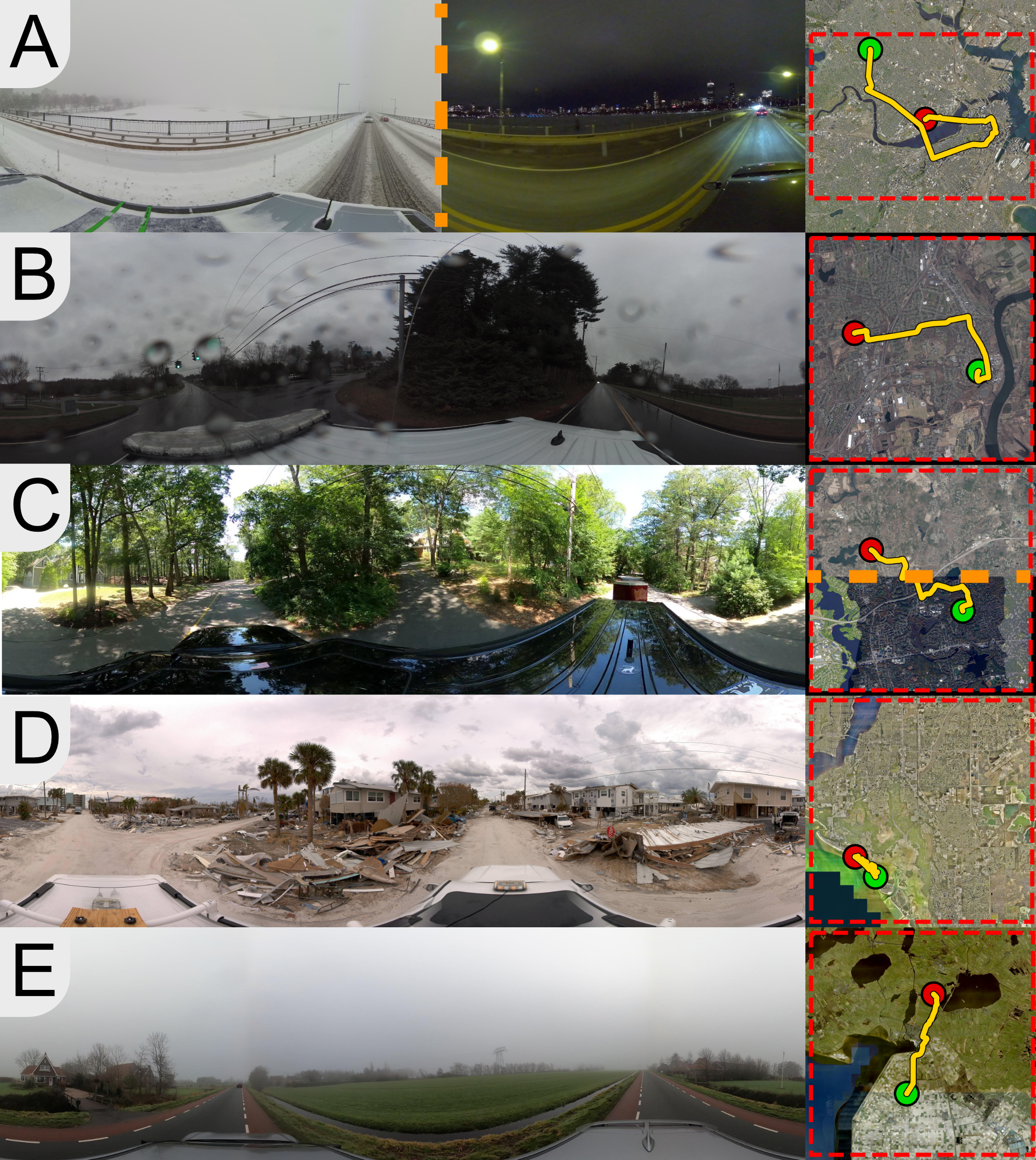}
    \caption{Representative panoramas, overhead map coverage (red dashed lines), and trajectories (yellow) for both the snowy and night Boston datasets (A), Middletown (B), Framingham (C), Fort Myers (D) and Noordoostpolder (E). The vehicle trajectory starts at green and ends at red. The trajectory driven for Boston was the same for both collects. Orange dashed lines separate the snowy and night panoramas in (A) and the consistent MassGIS mosaic above versus the Mixed-Sat Google mosaic below (discussed in Section \ref{sec:evaluation}) in (C).}
    \label{fig:created-datasets}
\end{figure}

Finally, we combine all of these sources of data together, pairing every panorama trajectory (or collection of unsequenced panoramas for the VIGOR cities) with overhead satellite imagery and OSM landmarks that cover a large region encompassing each respective set of panoramas. Sample panoramas, the full panorama trajectory, and overhead regions for some of the self-collected trajectories and Mapillary trajectories are shown in Figure \ref{fig:created-datasets}.

\subsection{Annotation Dataset}

\begin{keyidea}{How was the dataset extracted?}
To extract annotations from the panoramas, the procedure defined in Section \ref{sec:landmark_extraction} was followed using Gemini 3 Flash \cite{gemini3flash} with high thinking as the VLM with the prompt described in Table \ref{tab:prompt-structure}. The square pinhole reprojected images that were provided to the VLM had a $90^\circ$ vertical and horizontal FoV and a resolution of 1024\,px for pinhole images projected from VIGOR panoramas and 2048\,px for all other panoramas (as non-VIGOR panoramas were higher resolution). All prompts and scripts to perform this extraction are provided in the Github repository. Annotation extraction for the VIGOR cities (Chicago/Seattle/NewYork) consumed 808M/390M/32M input/thinking/output tokens, and all other panoramas consumed 116M/50M/3.1M input/thinking/output tokens. An example of annotations extracted from a panorama is shown in Figure \ref{fig:late_fusion_system}.
\end{keyidea}

\begin{keyidea}{How are we evaluating quality?}
To evaluate the quality of the extracted annotations, two human labelers independently inspected a random subset of 500 annotations per city in Chicago and Seattle (1,000 total). 
The labelers marked each annotation as valid (the landmark is visible somewhere in the panorama), valid with minor tag mistakes (defined as a tag mistake that is unlikely to affect its correspondence with the correct MSM landmark, e.g., identifying a building as 4 instead of 5 floors while getting the building name correct), or invalid (no such landmark is visible anywhere in the panorama).
During the inspection process, the labelers consulted with OSM and Google Maps for additional context.
Inter-rater agreement was substantial with Cohen's $\kappa$ = 0.70/0.75 \cite{cohen_coefficient_1960} on Chicago/Seattle, rising to  $\kappa= 0.72/0.8$ when ``minor'' is collapsed into ``valid''. The 87 of 1,000 annotations (8.7\%) on which the labelers disagreed were adjudicated by a third pass. The results in Table \ref{tab:human-eval} show the final annotation accuracy using the unified labels. Overall, \textasciitilde85\% of the evaluated annotations extracted by the VLM in Chicago and Seattle were valid or contained only minor tag mistakes.

\end{keyidea}

\subsection{Annotation Correspondence Dataset} \label{sec:landmark-correspondence-dataset}
\begin{keyidea}{Definition}
As part of this dataset, we also release the correspondence datasets used to train and validate the correspondence model described in Section \ref{sec:landmark_correspondence}. The annotation correspondence dataset is a set of $(\hat{a}^o, a^\M, m)$ tuples where $\hat{a}^o$ and $a^\M$ are a pair of annotations and $m$ denotes if $\hat{a}^o$ and $a^\M$ could describe the same landmark. To create the annotation correspondence training and validation dataset, we follow the procedure outlined in Section \ref{sec:landmark_correspondence} for Chicago and Seattle respectively. The prompt described in Table~\ref{tab:correspondence-prompt-structure} was used with Gemini 3 Flash with low thinking, consuming 42M/4.4M/5.4M input/thinking/output tokens total to produce 36K/40K/37K and 32K/27K/27K positive/easy negative/hard negative correspondences for Chicago and Seattle respectively. Some example correspondences can be seen in Figure \ref{fig:example_correspondences}.
\end{keyidea}

\begin{figure}
    \centering
    \definecolor{posfg}{HTML}{2E7D32}
\definecolor{negfg}{HTML}{C62828}
\definecolor{panofill}{HTML}{E3F2FD}
\definecolor{panodraw}{HTML}{1976D2}
\definecolor{osmfill}{HTML}{FFF3E0}
\definecolor{osmdraw}{HTML}{F57C00}

\newcommand{\boxtitle}[1]{{\normalfont\sffamily\bfseries #1}}

\newif\ifdebug \debugfalse

\begin{tikzpicture}[
    font=\sffamily\footnotesize,
    chip/.style={ 
        align=center,
        font=\sffamily, inner sep=0pt
    },
    chippos/.style={chip, text=posfg},
    chipneg/.style={chip, text=negfg},
    tagbox/.style={
        align=left,
        inner sep=0pt, font=\sffamily\footnotesize\ttfamily
    },
    panobox/.style={tagbox},
    osmbox/.style={tagbox},
    pmatch/.style={
        align=center, font=\sffamily, text width=3.1cm,
    },
    pmatchpos/.style={pmatch, text=posfg},
    pmatchneg/.style={pmatch, text=negfg}
]

\def\innersep{2pt}
\node[chippos, anchor=north west] (poslabel) at (0, 0) {%
    Positive Example $P(\mathrm{match}) = 0.997$
};

\node[panobox, anchor=north west] (pospano) at ($(poslabel.south west) + (0, -\innersep)$) {%
    \boxtitle{Set 1: Panorama Annotation}\\[2pt]
    amenity=restaurant\\
    name=Carnitas Don Pedro\\
    cuisine=mexican};

\node[osmbox, anchor=north east] (pososm) at ($(pospano.north west) + (\columnwidth, 0)$){%
    \boxtitle{Set 2: MSM Annotation}\\[2pt]
    name=Don Pedro Carnitas\\
    amenity=cafe\\
    addr:housenumber=1113\\
    addr:street=West 18th Street};

\node[fit=(pospano)(pososm), inner sep=0pt] (posrow) {};

\draw[gray!50] ($(pospano.west |- pososm.south) + (0, -\innersep)$) -- 
               ($(pososm.south east) + (0, -\innersep)$);

\node[chipneg, anchor=north west] at ($(pospano.west|- pososm.south) + (0pt, -2*\innersep)$) (neglabel) {%
    Negative Example
    $P(\mathrm{match}) = 0.591$
};

\node[panobox, anchor=north west] at ($(neglabel.south west) + (0pt, -\innersep)$) (negpano) {%
\boxtitle{Set 1: Panorama Annotation}\\[2pt]
amenity=hospital\\
name=UChicago Medicine\\
building=yes};

\node[osmbox, anchor=north east] at (pososm.east |- negpano.north) (negosm) {%
    \boxtitle{Set 2: MSM Annotation}\\[2pt]
    name=UIC College of Medicine\\
    building=yes\\
    addr:housenumber=1835\\
    addr:street=West Polk Street};

\node[fit=(negpano)(negosm), inner sep=0pt] (negrow) {};

\ifdebug
    \draw[red,thin] (0, 0)--(0, -200pt);
    \draw[red,thin] (\columnwidth, 0)--(\columnwidth, -200pt);
\fi

\end{tikzpicture}
    \caption{A positive and negative example from the annotation correspondence dataset. For the positive example, even though no key-value pair matches, these annotations describe the same physical landmark. The trained correspondence classifier assigns a true match probability of 0.997. For the negative example, even though both annotations are university-based and healthcare related, they are different physical landmarks. The trained correspondence classifier assigns a match probability of 0.591. This is below the decision point of 0.8 and is therefore treated as an invalid match.}
    \label{fig:example_correspondences}
\end{figure}

\begin{keyidea}{How are we evaluating quality?}
To evaluate the quality of the correspondence dataset, two labelers independently rated a random subset of 1,000 proposed correspondence pairs per city in Chicago and Seattle. Each pair consisted of one panorama-extracted annotation and one OSM landmark; the labelers were shown only the two sets of annotations, without the dataset's proposed label or difficulty class, and marked each pair as the same landmark or different. Pairs were sampled equally from the dataset's proposed positives and proposed negatives, with the latter drawn proportionally across easy and hard difficulty levels. Inter-rater agreement was high with $\kappa = 0.88/0.90$ for Chicago/Seattle. The 112 of 2,000 pairs (5.6\%) on which the labelers disagreed were adjudicated by a third pass, and the metrics comparing the VLM's correspondence labels to the human consensus labels are reported alongside the extraction evaluation in Table~\ref{tab:human-eval}. Finally, with these datasets, we now turn to using them to evaluate \name{}.
\end{keyidea}

\begin{table}[t]
  \centering
  \caption{Human evaluation of the panorama annotation extractor and the
  annotation correspondence dataset}
  \label{tab:human-eval}
  \begin{tabular}{lccc}
  \toprule
                                              & Chicago & Seattle & Combined \\
  \midrule
  \multicolumn{4}{l}{\textit{Panorama annotation extraction}
                     ($n{=}500\,/\,500\,/\,1000$)} \\[2pt]
  \quad Valid                               & 83.6\% & 78.2\% & 80.9\% \\
  \quad Valid, minor tag mistakes           &  4.2\% &  5.0\% &  4.6\% \\
  \quad Invalid          & 12.2\% & 16.8\% & 14.5\% \\
  \midrule
  \multicolumn{4}{l}{\textit{Annotation correspondence}
                     ($n{=}1000\,/\,1000\,/\,2000$)} \\[2pt]
  \quad Precision                             & 0.976  & 0.976  & 0.976  \\
  \quad Recall                                & 0.859  & 0.850  & 0.855  \\
  \quad F1                                    & 0.914  & 0.909  & 0.911  \\
  \bottomrule
  \end{tabular}
  \end{table}

\section{Evaluation} \label{sec:evaluation}
\begin{keyidea}{To evaluate our method, we aim to provide evidence for the following questions:
    \begin{itemize}
        \item Evaluation with simulated trajectories in urban environments using imagery from the VIGOR Dataset?
        \item How does our approach/use of semantics fair across a wide range of settings? Evaluation in snowy, low light, disaster, rural settings etc.
    \end{itemize}
    }
In our evaluation, we show that \name{} better enables robust and generalizable localization against prior overhead data across a diverse set of real-world data. We first describe in detail our experimental setup, baseline approaches, and evaluation infrastructure that enables us to evaluate each approach.
\end{keyidea}

\subsection{Baselines and Ablations}
\begin{keyidea}{We implement WAG with DINO features and a histogram filter. We train it on Chicago}
\textbf{WAG} \cite{downes_city-wide_2022} performs city-scale cross-view localization by matching learned features between street-level panoramas and overhead satellite imagery. We substituted the original VGG-16 backbone with frozen DINOv3 \cite{simeoni2025dinov3} patch level features, which we found substantially improved retrieval performance over the original architecture. 
WAG serves as our primary cross-view localization baseline as it operates without restrictions over large overhead areas.
\end{keyidea}

\begin{keyidea}{Train WAG on Chicago VIGOR data}
The siamese network for the WAG baseline was trained on all panoramas in Chicago. A batch size of 18 was used for 60 epochs with the AdamW optimizer and a learning rate of $0.0001$. All satellite patches were resized to $320\times320$\,px and panoramas were resized to $640\times320$\,px. Hard negative mining was enabled after epoch 20. A pairwise contrastive loss was used with weights of 5 for positive pairs and 20 for negative pairs. Semi-positive panorama-satellite matches (where the panorama lies outside of the middle 50\% of the satellite patch) were treated identically to positives (within the center 50\%), differing from Downes \textit{et al.}~\cite{downes_city-wide_2022}. Training took 5\,h on an Nvidia GeForce RTX 5090. To select a variance for the Gaussian observation likelihood used in Downes \textit{et al.}~\cite{downes_city-wide_2022} (shown in Equation \ref{eq:obs_liklihood_wag}), we found its maximum likelihood fit on true panorama-satellite patch pairs for the Seattle VIGOR dataset resulting in $\sigma_\text{WAG} = 0.1809$. This same fitting method was used for all approaches described below. Note that Equation \ref{eq:obs_liklihood_wag} is similar to Equation \ref{eq:obs_liklihood_landmark} but uses a different form of the residual as the similarity values $s_{o,i}$ produced by WAG are cosine similarities and bounded in magnitude. 

\end{keyidea}
\begin{equation} \label{eq:obs_liklihood_wag}
p(o\mid x_i, \I) \propto \exp{\frac{-1}{\sigma_\text{WAG}^2} \left( \max_j s_{o, j} - s_{o,i} \right)^2}
\end{equation}

\begin{keyidea}{We compare against a fused OSM tile approach as it closely matches other baselines while relaxing their assumptions}
\textbf{WAG + OSM} represents approaches that keep only coarse semantics and drop instance details (e.g., shop names or house numbers) from OSM and is based on Zhou \textit{et al.}~\cite{zhou_efficient_2021}. This approach is very similar to WAG, but instead of using satellite images as the overhead data, Zhou \textit{et al.}~\cite{zhou_efficient_2021} uses rendered OSM tiles as the overhead data representation, claiming that the semantics in the OSM tiles provide more invariance to changing conditions. To align this approach to our problem setting we made a few changes. First, as we assume heading information is available, we removed the heading estimation from Zhou \textit{et al.}~\cite{zhou_efficient_2021}. Second, we removed the strong prior that the only feasible locations are on roads present in the OSM data.
Third, we reduced the sampling density of overhead tiles from 1\,m (as done in Zhou \textit{et al.}~\cite{zhou_efficient_2021}) to 36\,m as is provided by our datasets and our approach to make scaling to the large environments more achievable.
Finally, to make the comparison fair, we fused the outputs of Zhou \textit{et al.}~\cite{zhou_efficient_2021} with WAG in the same manner we describe in Section \ref{sec:approach}. We observed this fusion to generally improve metrics for this baseline across evaluation datasets. This fusion narrows the key difference between Zhou \textit{et al.}~\cite{zhou_efficient_2021} and our approach to the rasterization of OSM tiles instead of treating landmarks independently. The OSM component of this baseline was trained with the same hyperparameters described above, and its Gaussian observation likelihood (matching Equation \ref{eq:obs_liklihood_wag}) $\sigma_\text{OSM} = 0.2462$ was also fit on Seattle identically to WAG.
\end{keyidea}

\begin{keyidea}{We also compare against an early-fusion based approach}
\textbf{\name{}-EF} is an alternative ``early fusion''  approach that fuses landmark information with image information early instead of estimating both independently and fusing their distributions ``late'' with product of experts. \name{}-EF adds a transformer encoder to the end of each panorama/satellite model. This transformer takes in the frozen WAG produced embedding, a CLS token, and self-attends both with panorama/satellite landmark tokens produced by the annotation encoder in Figure \ref{fig:landmark_correspondence_model}. The CLS token is then normalized and used as the panorama/satellite embedding. An InfoNCE contrastive loss \cite{oord2018representation} is combined with a distillation loss that rewards high cosine similarity between model outputs and the frozen WAG embeddings to encourage the model to only deviate from WAG's predictions when the landmark tokens are discriminative. \name{}-EF again was trained only on Chicago and used the Gaussian observation likelihood (Equation \ref{eq:obs_liklihood_wag}) with a fit $\sigma_\text{EF} = 0.1876$ on Seattle.
  
\end{keyidea}

\subsection{Approach Details} \label{sec:evaluation_details}
\begin{keyidea}{Correspondence model training}
The correspondence model described in Section \ref{sec:landmark_correspondence} was trained with a batch size of 512 for 20 epochs and a learning rate of $3\times10^{-4}$ with a binary cross entropy loss. We used the correspondences generated from the Chicago subset as the training set and those from Seattle served as the validation set. This achieved a train and validation ROC-AUC score of 0.993 and 0.976 respectively after training for 90\,s on an Nvidia GeForce RTX 5090.

With this correspondence model we generated confidences that an individual panorama landmark corresponded to an OSM landmark. We followed the procedure outlined in Section \ref{sec:landmark_observation_likelihood} with a Hungarian matching threshold of 0.8. With a final landmark-based similarity score for each panorama-overhead patch pair, we calculated $\sigma_{l}$ in Equation \ref{eq:obs_liklihood_landmark} using the same MLE fit as done for the baseline approaches, except we exclude positive matches where $s_{o,i}=0$ which otherwise lead to a poor fit. Performing this fit on Seattle resulted in $\sigma_l=0.4673$.

\end{keyidea}

\begin{keyidea}{Runtime Statistics, argue for realtime deployment}
The pipeline processing time was measured for a subset of the locations in the Chicago dataset.
The median processing time was 36.6\,s, with landmark extraction taking 32.9\,s and correspondence computation taking 3.7\,s.
Assuming a typical walking speed of 2\,m/s of a quadrupedal robot, this translates to roughly 75\,m of travel between processed observations.
We additionally measured the cost of replacing the learned correspondence model with a local language model prompted to match a panorama's landmarks directly against the OSM landmarks for Chicago. As the 28{,}582 sets of unique annotations in Chicago's landmarks far exceed the models' context window, we streamed them through in chunks and timed inference on an Nvidia GeForce RTX 5090. This yielded roughly 5.4/7.6\,min per panorama without/with thinking for Gemma 3 27B \cite{gemma_2025_gemma3}, roughly two orders of magnitude slower than the 3.7\,s learned correspondence step.

\end{keyidea}

\subsection{Path Experiment Setup}
\begin{keyidea}{Path sampling}
To evaluate localization performance, we need robot trajectories. As the VIGOR panoramas were sampled from a large set of Google Streetview panoramas, there is not a natural trajectory through the panoramas. To obtain an evaluation set of trajectories, we generated a set of 5000 5\,km trajectories using walks through a graph where nearby panoramas are connected to each other. 

The self-collected and Mapillary datasets were each collected as a single sequence of panoramas. Panoramas were spatially downsampled with a minimum spacing of 5\,m, providing the number of panoramas per trajectory reported in Table \ref{tab:dataset-stats}.
We subsampled the full trajectory into overlapping continuous 3\,km segments spread uniformly along the full trajectory's length, sampling 500 forward and 500 backwards.  %
\end{keyidea}

\begin{keyidea}{Rest of setup}
We evaluated each approach on this set of evaluation trajectories.
For all approaches we used a histogram filter to instantiate the Bayes filter.
With \textasciitilde20\,m$\times$20\,m cell size we observed that using a histogram filter was effective at representing the state probability density over large city-scale areas while allowing us to avoid particle starvation issues and tuning of additional hyperparameters as compared to a particle filter.
One limitation of a histogram filter is a lower bound on achievable error, but \textasciitilde20\,m$\times$20\,m grid cells were sufficient for our localization task. 
Odometry $u_t\in\mathbb{R}^2$ was derived by taking the deltas in latitude and longitude between sequential panoramas in the evaluation path.
Odometry noise was sampled from $\mathcal{N}(0,\sigma_n^2d_{ij})$, where $\sigma_{n}$ is fixed at 0.141\,m/${\sqrt{\text{m}}}$ and $d_{ij}$ is the distance between the $i$th and $j$th panorama in the trajectory in meters, and was added to the odometry measurements at each step, which corresponds to 1\,m error standard deviation for 50\,m of travel. 

The prediction step of the histogram filter shifts the current distribution by the odometry and then applies Gaussian smoothing over the 2D histogram similarly with a variance of $\sigma_{np}^2 = 0.141^2 d_{ij}$.
\end{keyidea}
\begin{keyidea}{Metrics}
\,We report two types of metrics for each trajectory evaluation. \textbf{Final Error (m)} measures the Euclidean distance between the estimated position at the end of the trajectory and ground truth, capturing the final accuracy of each approach. We used the mean of the histogram filter as the estimated location for this metric. \textbf{Expected Distance to Convergence} (EDC) at $\delta$ threshold (m), detailed in Equation \ref{eq:expected_distance_to_convergence}, captures how much distance is traveled weighted by the probability of the belief being at least $\delta$ away from the true location. The faster the belief converges to the true location, the lower this metric becomes. We report both of these metrics averaged across all trajectories $\tau \in \mathcal{T}$ in an evaluation set.
\begin{equation}
\text{EDC}_{\delta}(\tau) = \int_{\tau}p(||x(s) - x^*(s)||_2 > \delta) ds
\label{eq:expected_distance_to_convergence}
\end{equation}
\end{keyidea}

\subsection{Path Experiment Results}

\begin{figure*}
    \centering
    \includegraphics[width=\textwidth]{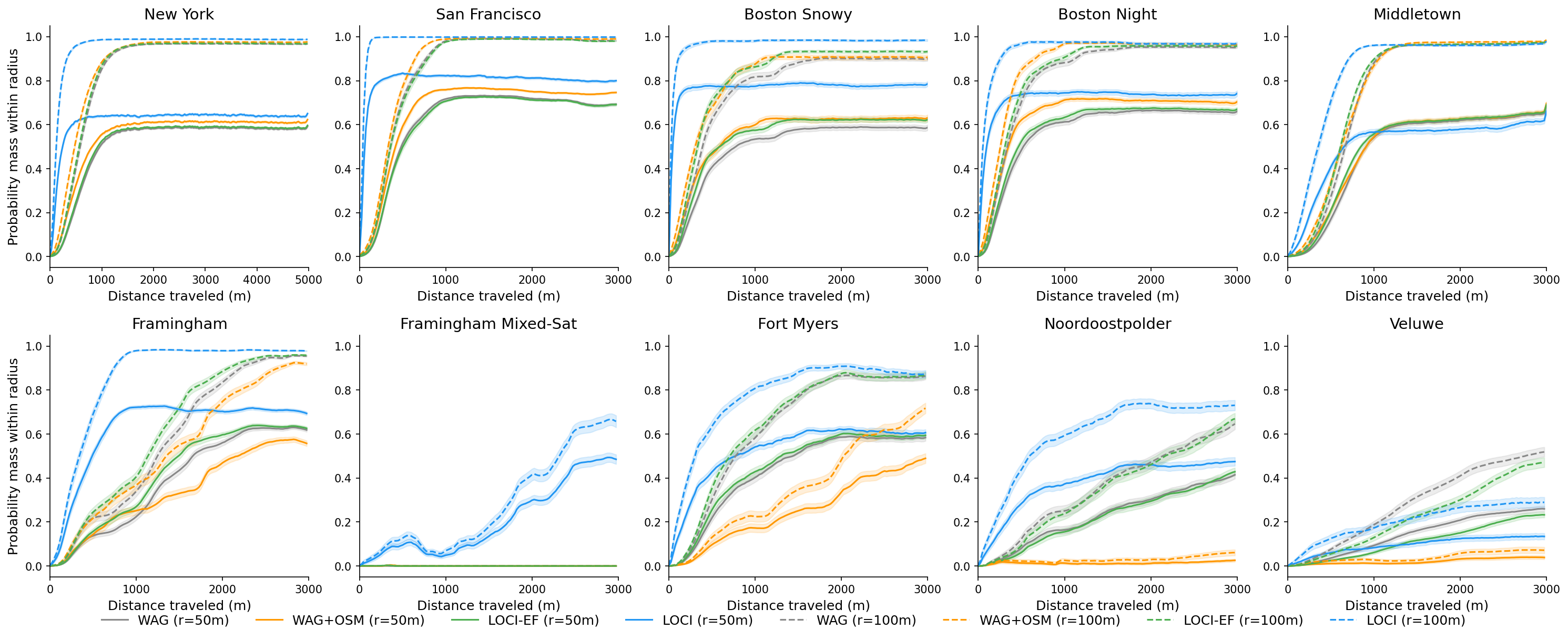}
    \caption{Probability mass within $r=50$\,m and $r=100$\,m of the true location for all evaluation environments. Shaded regions show $95\%$ CI for the mean (using standard error of the mean). Across all environments but Veluwe, on average with high confidence \name{} places more probability mass closer to the true location faster than other approaches. This demonstrates that VLM-extracted semantic landmarks and MSMs can generally improve cross-view localization across very different test environments.}
    \label{fig:convergence-curves}
\end{figure*}

\begin{table*}[t]
  \centering
  \caption{Expected distance to convergence (m) and final localization error (m) across environments. Values shown as mean $\pm$ 1.96 $\cdot$ SEM (normal-approximation 95\% CI for the mean). \textbf{Bold} indicates the best method. Evaluation trajectories were 3\,km for all self-collected and Mapillary trajectories, and 5\,km for New York.}
  \label{table:convergence-and-final-error}
  \begin{tabular}{lrrrrrrrr}
  \toprule
   & \multicolumn{4}{c}{Mean EDC$_{100}$ (m) $\downarrow$} & \multicolumn{4}{c}{Mean Final Error (m) $\downarrow$} \\
  \cmidrule(lr){2-5} \cmidrule(lr){6-9}
   & WAG & WAG+OSM & LOCI-EF & LOCI & WAG & WAG+OSM & LOCI-EF & LOCI \\
  \midrule
  New York            & 718.5$\pm$7.0            & 606.3$\pm$8.0   & 691.4$\pm$6.8   & \textbf{184.5$\pm$3.5}   & 9.2$\pm$0.1              & 9.6$\pm$0.2           & \textbf{9.1$\pm$0.1}  & 11.2$\pm$0.2              \\
  San Francisco        & 428.3$\pm$9.4            & 371.2$\pm$7.2   & 436.7$\pm$9.8   & \textbf{39.9$\pm$1.1}    & 10.6$\pm$0.5             & \textbf{9.2$\pm$0.2}  & 10.4$\pm$0.5          & 11.0$\pm$0.4              \\
  Boston Snowy         & 706.8$\pm$22.1           & 588.3$\pm$16.2  & 566.1$\pm$17.3  & \textbf{92.3$\pm$4.8}    & 32.8$\pm$1.6             & 33.2$\pm$2.2          & 27.6$\pm$1.1          & \textbf{22.0$\pm$0.9}     \\
  Boston Night         & 519.4$\pm$16.3           & 363.2$\pm$11.0  & 468.4$\pm$14.0  & \textbf{160.4$\pm$5.6}   & 23.7$\pm$0.9             & \textbf{22.9$\pm$0.8} & 23.2$\pm$0.9          & 24.6$\pm$1.2              \\
  Middletown           & 717.2$\pm$10.5           & 674.4$\pm$13.0  & 674.7$\pm$10.4  & \textbf{443.7$\pm$10.4}  & 17.7$\pm$0.5             & 20.6$\pm$0.6          & \textbf{17.0$\pm$0.5} & 24.6$\pm$0.6              \\
  Framingham           & 1272.0$\pm$28.4          & 1393.3$\pm$39.5 & 1143.4$\pm$27.4 & \textbf{418.7$\pm$12.4}  & 19.9$\pm$0.8             & 26.5$\pm$0.9          & 18.1$\pm$0.7          & \textbf{15.4$\pm$0.7}     \\
  Fram. Mix-Sat & 3007.2$\pm$0.3           & 3006.8$\pm$0.3  & 3006.9$\pm$0.3  & \textbf{2146.3$\pm$36.1} & 3393.9$\pm$41.9          & 3095.3$\pm$38.4       & 3360.5$\pm$41.6       & \textbf{1349.7$\pm$111.2} \\

  Fort Myers           & 1070.1$\pm$36.5          & 1929.3$\pm$46.9 & 1022.2$\pm$37.6 & \textbf{642.6$\pm$24.9}  & 96.8$\pm$21.2            & 960.5$\pm$170.3       & 89.0$\pm$18.7         & \textbf{33.2$\pm$2.9}     \\
  Noordoostp.      & 1937.2$\pm$52.5          & 2916.3$\pm$12.5 & 1969.2$\pm$50.5 & \textbf{1202.3$\pm$43.9} & 2727.2$\pm$274.5         & 7983.5$\pm$291.2      & 2302.4$\pm$259.9      & \textbf{128.3$\pm$12.3}   \\
  Veluwe               & \textbf{2137.9$\pm$37.4} & 2883.0$\pm$18.5 & 2319.0$\pm$34.2 & 2387.3$\pm$40.0          & \textbf{858.2$\pm$104.4} & 5845.1$\pm$284.7      & 1263.9$\pm$216.1      & 2617.3$\pm$270.5          \\
  \bottomrule
  \end{tabular}
\end{table*}

\begin{keyidea}{Table and plots}
Aggregate statistics across paths for \name{} and all baselines can be seen in Table \ref{table:convergence-and-final-error} and probability mass within $r=50$ and $r=100$\,m from the true position as a function of distance traveled can be seen in Figure \ref{fig:convergence-curves}. %
Veluwe aside, \name{} has marginally higher mean final error (by at most 8 meters, within a single histogram filter cell), while placing more probability mass closer to the true location faster than all other approaches, significantly reducing EDC$_{100}$ scores. 
\end{keyidea}

In the city environments (left four plots on the top row of Figure \ref{fig:convergence-curves}) all approaches converged to within 100\,m of the true location by the end of the trajectory, except for on the Boston Snowy dataset where only \name{} regularly accumulated all probability mass within 100\,m. In these landmark-rich environments, we saw \name{} converged to the true location much faster (within 100-500\,m) compared to all other approaches (900+\,m depending on the dataset). Among the baseline approaches, we saw rasterized OSM helping WAG + OSM in Boston Night and New York, and \name{}-EF offered limited benefit over just WAG in most city environments. Notably, \name{}-EF significantly under-performing \name{} suggests that the approach fails to combine image and landmark data in a way that generalizes as well past its training data. We see a similar pattern in Middletown, a more suburban dataset captured during a rainstorm. As landmarks are less dense in Middletown than in Boston Snowy, \name{} did not localize as quickly as in the city datasets, but still converged faster than other approaches.

Framingham is a self-similar suburban environment, and we use it to probe how much each approach relies on the quality of the overhead imagery. Holding all else fixed, we evaluate against two satellite sources: the single-date MassGIS spring 2025 orthomosaic mentioned in Section \ref{sec:dataset}, and a second mosaic (``Framingham Mixed-Sat'') stitched from tiles pulled from Google Maps. Though all tiles were pulled on the same date, the satellite tiles themselves come from different imaging dates, which leaves visible seams between the west and east regions (leaf-on data) and the rest of the patches (leaf-off data). See the overhead image in the bottom of Figure \ref{fig:created-datasets}-C. Such temporally inconsistent mosaics are common in practice. Because only the overhead imagery differs between the two conditions, the comparison isolates each approach's reliance on consistent satellite appearance. With the clean overhead, all approaches converge to within 100\,m by the end of the 3\,km trajectories, though \name{} converges roughly three times faster than WAG (EDC$_{100}$ of 418.7\,m versus 1272.0\,m), leveraging house numbers and street names to disambiguate otherwise repetitive suburban blocks. On Framingham Mixed-Sat, all approaches except \name{} fail to gain any traction (EDC$_{100}\approx3007$\,m, near the trajectory-length ceiling), as the satellite-appearance matching consistently biased approaches towards the leaf-on imagery on the east/west side of the overhead region. \name{} does not fully localize (1349.7\,m final error), but retains more localization signal than any other approach with the help of instance-level semantics.

The Fort Myers, Noordoostpolder, and Veluwe datasets cover a much larger area (627\,km$^2$ instead of \textasciitilde50\,km$^2$). As we started each trial with a uniform prior over the dataset area, this larger area can make localization more difficult. Evaluation on the Fort Myers dataset explored how robust approaches were to severe changes in environment appearance, with sand and debris covering many of the roadways and vehicles and buildings overturned. \name{} placed probability mass close to the true position meaningfully faster than all other approaches. 

In Noordoostpolder we see slower and less successful localization across approaches, with \name{} still leading with an EDC$_{100}$ of 1202.3 compared to the next best score of 1937.2 from WAG. Interestingly, WAG+OSM performed quite poorly, suggesting that the rasterized OSM tiles were more difficult to match against than the satellite imagery in these more landmark-sparse environments. 

Veluwe was an environment where \name{} was outperformed by WAG. Veluwe's trajectory lies on a landmark-sparse stretch of multi-lane highway. Though Framingham/Middletown/Fort Myers/Noordoostpolder had similar landmark per panorama density, their panorama landmarks were regularly more discriminative (street name signs, house numbers). We observed some landmarks in Veluwe, like communication towers, were inconsistently mapped in OSM, pulling probability mass away from the true location towards where the landmarks were mapped. Veluwe's key difference was that landmarks were dominantly mile markers, overpasses, and signs referring to landmarks far from the robot's current location, violating the proximity assumption we make in our observation model. We discuss possible ways to address this in Section \ref{sec:future_work}.

\begin{keyidea}{Takeaway: adding landmarks speeds up convergence. In self-similar environments without distinctive overhead features, or when image inputs are drastically different than expected, landmarks add robustness to image-only approaches that allows them to localize more accurately.}
Across most environments, incorporating landmarks meaningfully accelerated convergence, consistent with the hypothesis that semantic landmarks provide a localization signal complementary to image-based retrieval. The gains were largest where either image-only matching was most strained or landmarks were dense and informative. In landmark-dense cities, where overhead imagery was informative but self-similar across blocks, \name{} converged 3-10 times faster than WAG while reaching comparable final accuracy. In Fort Myers and Framingham, where domain shift and mixed satellite image dates degrade the overhead signal, landmarks additionally drove faster and more accurate localization. WAG+OSM, which retains only coarse semantic classes through rasterization, offered limited benefit and was actively harmful in Fort Myers, Noordoostpolder, and Veluwe, where the rasterized priors did not survive the appearance shift or the self-similarity in landmark-sparse settings. This contrast supports that \emph{instance-level} semantics like street names, house numbers, and named landmarks (rather than coarse category maps) can disambiguate self-similar regions and resist visual domain shift more effectively in many environments. Finally, \name{}-EF's failure to match \name{} across most environments indicates that access to both modalities is not sufficient; how landmark and image observations are represented and combined affects whether the filter generalizes beyond its training distribution.
\end{keyidea}

\section{Related Work} \label{sec:related_works}

\begin{keyidea}{Image-based cross-view localization — datasets, CNNs, transformers, WAG particle filter}
The problem of image-based cross-view localization has seen significant progress as datasets like CVUSA \cite{cvusa}, VIGOR \cite{zhu_vigor_2021}, and CV-Cities \cite{huangCVCities2024} have enabled deep learning approaches. Early methods leveraged convolutional neural networks \cite{tian2017cross} and siamese architectures \cite{hu2020image} to learn correspondences between overhead satellite images and egocentric observations, while more recent work has adopted transformers \cite{zhu_transgeo_2022, yuan_cross-attention_2024} to improve cross-view matching. Relevant to our work,  WAG \cite{downes_city-wide_2022} uses a siamese network to learn embeddings for ego panoramas and satellite images, using the learned similarity as an observation likelihood in a particle filter. WAG demonstrates convergence to the correct position starting from a city-scale location prior after approximately 20 km of travel. Though modern DINOv3 features greatly improve WAG's speed of convergence to a kilometer or two, image-based cross-view localization approaches miss out on readily available, and often highly discriminative semantic information that is not captured in satellite imagery, like business or street names, because of the perspective transform.

\end{keyidea}

\begin{keyidea}{OSM-based methods — rasterization loses semantics; no egocentric semantic extraction}
Metric-Semantic maps like OpenStreetMap (OSM) capture this kind of data, and provide cross-view localization approaches with a way to leverage this otherwise lost semantic information. Samano \textit{et al.} \cite{samano2020you} match rasterized OSM maps with egocentric panoramas by learning a shared embedding space. OrienterNet \cite{sarlin_orienternet_2023} learns to project egocentric images into a bird's-eye-view (BEV) representation, which is then aligned with an encoding of OSM data; however, it assumes a position prior. Zhou \textit{et al.} \cite{zhou_efficient_2021} integrate OSM maps and egocentric panoramas into a particle-filter-based localization framework. Hu \textit{et al.} \cite{hu2024combining} extend this line of work by fusing both satellite imagery and OSM data for the overhead representation, demonstrating improved robustness over single-modality approaches. A limitation shared by these methods is that they rasterize OSM data, discarding much of the rich semantic information available, such as landmark descriptions and place types.  These approaches also do not explicitly extract semantic information from the egocentric images.
\end{keyidea}

\begin{keyidea}{CrossText2Loc — VLM semantics but single-modality retrieval, tight prior, no sequential localization}
CrossText2Loc \cite{ye_where_2025} extracts semantic descriptions from egocentric images using a VLM. However, CrossText2Loc retrieves against satellite imagery \emph{or} OSM data independently rather than fusing both modalities. Though CrossText2Loc still rasterizes its OSM data, it includes the names of some businesses in that raster, allowing it to match egocentric semantics with overhead semantics. It also assumes a tight location prior, matching against only the 100 nearest overhead candidates to the true location, whereas our approach assumes only a city-scale location prior spanning hundreds of square kilometers. Furthermore, CrossText2Loc discards the egocentric image after extracting semantics from it, provides the VLM with privileged information during landmark extraction, and does not focus on sequential localization. In \name{}, we leverage satellite data, OSM data, egocentric panoramas, and semantics extracted from those panoramas. We richly encode the semantic content in both OSM data and egocentric observations and provide no privileged information. 
\end{keyidea}

\begin{keyidea}{Text-based SLAM and place recognition — related but don't solve general localization}
Other works outside of cross-view localization have investigated using text observed by the robot as an invariant feature for simultaneous localization and mapping \cite{li_textslam_2023}. Text extraction ``in the wild'' poses challenges that other works address jointly with place recognition \cite{raisi2022text}. In \cite{mti6110102}, the authors match egocentric images and extracted text to ``places of interest'' from an OSM database. These approaches do not address the same cross-view localization problem as \name{}.
\end{keyidea}

\section{Limitations and Future Work} \label{sec:future_work}
\begin{keyidea}{Panoramas can often see beyond current patch.}
One key opportunity for improvement we encountered with \name{} in Veluwe is better handling of landmark visibility. Though prompted to only extract proximate landmarks, frequently VLMs would return notable but distant semantic landmarks, such as the name of a nearby town on a highway sign, a church a few hundred meters away, or a notable skyscraper like Willis Tower on the Chicago horizon. Proper identification and integration of these features into the observation model could greatly improve localization (e.g., a known bearing to Willis Tower would greatly narrow your belief). 
\end{keyidea}

\begin{keyidea}{Lack of uncertainty from VLM observations}
Another limitation we observed was the lack of calibrated uncertainty available from the VLM-extracted panorama semantic landmarks. This lack of uncertainty makes \name{} more vulnerable to hallucinations. One interesting direction of future work is the filtering of extracted landmarks over time before providing the set of landmarks to \name{} to calculate the updated observation likelihood. For example, if panoramas taken every few meters were each used for landmark extraction, corresponding landmarks could be inspected and vetted for robustness to approximate which properties of these landmarks are reliable and not. As an example, we often saw that street name and house number label extraction improved as the street sign got closer in the panorama.
Finally, all of our localization results only exercise the cloud-hosted Gemini 3 Flash model, which would require network connectivity during deployment. It is unclear how well this approach transfers to the rapidly developing model families that are practical to run onboard. 
\end{keyidea}

\section{Conclusion}
In this work, we introduced \textit{\name{}}, a cross-view localization framework that integrates information from OSM to improve localization accuracy and convergence speed. By leveraging both satellite imagery and rich semantic and textual annotations from OSM, our approach mitigates the challenges posed by drastic perspective differences between egocentric and overhead views. Through extensive evaluations on real-world data we demonstrate that incorporating richer OSM annotations significantly enhances localization performance, reducing the trajectory length required for accurate localization convergence when compared to previous approaches. 

Our experiments show that \name{}, trained only in Chicago and validated only in Seattle, generalizes well to new locations and weather conditions, and outperformed image-only and other OSM-based methods in most urban and rural environments. \name{} provides a step towards rapid localization against prior spatial information, enabling robots to operate with richer context in GPS-denied environments.

\section*{Acknowledgments}
This work was supported in part by ARL under Grants W911NF-21-2-0150 and W911NF-17-2-0181. The views and conclusions contained in this document are those of the authors and should not be interpreted as representing the official policies, either expressed or implied, of the Army Research Office or the U.S. Government. The U.S. Government is authorized to reproduce and distribute reprints for Government purposes notwithstanding any copyright notation herein. The authors gratefully acknowledge Lambda Labs and Tensorpool for providing computational resources for this work.

\bibliographystyle{IEEEtran}
\bibliography{zotero_references, other_references}  %

\end{document}